\pdfoutput=1 

\documentclass[12pt]{amsart}
\usepackage[utf8]{inputenc}
\usepackage[T1]{fontenc}

\usepackage[a4paper, left=2.7cm, right=2.7cm, bottom=2.5cm]{geometry}
\usepackage[foot]{amsaddr}

\usepackage{amssymb,amsmath,amstext,amsfonts,amsthm,braket}
\usepackage{mathtools}
\usepackage{hyperref}
\usepackage{tikz}
\usetikzlibrary{decorations.pathreplacing}
\usepackage{nicematrix}
\usepackage{graphics}
\usepackage[misc]{ifsym}
\usepackage{bbding}
\usepackage{todonotes}
\usepackage{blkarray}
\usepackage{paralist}
\usepackage{graphicx}
\usepackage{booktabs}
\usepackage{xcolor}
\usepackage{siunitx}
\usepackage{multirow}
\usepackage[activate={true,nocompatibility},final,tracking=true,kerning=true,spacing=true,factor=1100,stretch=10,shrink=10]{microtype}
\usepackage{dsfont}
\usepackage{resizegather}
\usepackage{todonotes}
\usepackage{multicol}
\usepackage{tabularx}

\usepackage{algorithm}
\usepackage{algorithmic}

\usepackage{mathrsfs}  
\hypersetup{
    colorlinks=true,
    linkcolor=orange!80!black,
    filecolor=magenta,      
    urlcolor=teal,
    citecolor=teal,
}

\definecolor{myblue}{rgb}{0.00000,0.44700,0.74100}
\definecolor{myorange}{rgb}{0.8500, 0.3250, 0.0980}
\definecolor{myyellow}{rgb}{0.9290, 0.6940, 0.1250}
\definecolor{mypurple}{rgb}{0.4940, 0.1840, 0.5560}
\definecolor{mygreen}{rgb}{0.4660, 0.6740, 0.1880}

\usepackage{listings}

\DeclareMathOperator{\argmin}{argmin}
\DeclareMathOperator{\sgn}{sgn}

\newcommand{\R}{\mathbb{R}}
\newcommand{\Z}{\mathbb{Z}}
\newcommand{\A}{\mathcal{P}}
\newcommand{\T}{\mathcal{T}}
\newcommand{\id}{\mathds{1}}

\newtheorem*{theorem*}{Theorem}
\newtheorem{theorem}{Theorem}
\newtheorem{observation}[theorem]{Observation}
\newtheorem{conjecture}[theorem]{Conjecture}
\newtheorem{proposition}[theorem]{Proposition}
\newtheorem{lemma}[theorem]{Lemma}

\theoremstyle{remark}
\newtheorem{remark}[theorem]{Remark}

\theoremstyle{definition}
\newtheorem{definition}[theorem]{Definition}

\makeatletter
\newcommand{\neutralize}[1]{\expandafter\let\csname c@#1\endcsname\count@}
\makeatother

\newenvironment{thmbis}[1]
  {\renewcommand{\thetheorem}{\ref{#1}$'$}%
   \neutralize{theorem}\phantomsection
   \begin{theorem}}
  {\end{theorem}}

\begin{document}

\title{Pathological Regularization Regimes in Classification Tasks}

\author{Maximilian Wiesmann}
\address{Max Planck Institute for Mathematics in the Sciences, Inselstr.\ 22, 04103 Leipzig, Germany}
\email{wiesmann@mis.mpg.de}

\author{Paul Larsen}
\address{Fodoj GmbH, Prinzregentenstrasse 54, 80538 Munich, Germany}
\email{paul@mkdev.me}

\begin{abstract}
    In this paper we demonstrate the possibility of a trend reversal in binary classification tasks between the dataset and a classification score obtained from a trained model. This trend reversal occurs for certain choices of the regularization parameter for model training, namely, if the parameter is contained in what we call the \emph{pathological regularization regime}. For ridge regression, we give necessary and sufficient algebraic conditions on the dataset for the existence of a pathological regularization regime. Moreover, our results provide a data science practitioner with a hands-on tool to avoid hyperparameter choices suffering from trend reversal. We furthermore present numerical results on pathological regularization regimes for logistic regression. Finally, we draw connections to datasets exhibiting Simpson's paradox, providing a natural source of pathological datasets.

\end{abstract}

\maketitle

\section{Introduction}

While adversarial datasets are hand-crafted examples designed to trick a machine learning algorithm (see for example in \cite{goodfellow2014explaining}), naturally occurring datasets can similarly confuse these algorithms. These datasets would affect unsuspecting data scientists with the bad luck to be given a prediction or classification task for such a dataset. Rather than an attack by an adversary, such datasets would more closely resemble diseases that can be avoided or whose impact can be mitigated, hence we refer to them as ``pathological'' datasets.

In this paper, we demonstrate a family of binary classification random processes that are pathological for ridge and logistic regression. For ridge regression, we give precise, algebraic conditions for a dataset to be pathological. More precisely, for any three-dimensional binary dataset, we give an exact range of regularization hyperparameters that result in the fitted classification algorithm getting the data trends exactly wrong.

Datasets showing this pathological behavior occur with positive probability. Moreover, there exists an intriguing connection to Simpson's paradox: the proportion of pathological datasets among those exhibiting Simpson's paradox is particularly high. In the case of logistic regression we even conjecture that all Simpson datasets are pathological.

It should be noted that Simpson's paradox does occur in nature, for example in university admissions \cite{Bickel398}, clinical trials \cite{ABRAMSON19921480} and death-penalty judgments \cite{norton2015simpson}.  Note that these domains---access to education, medical research and judicial decisions---all fall under ``high risk AI'' as defined by the EU's AI Act \cite{euaiact2021}.

\subsection{Illustrating Example}
\label{subsec:Example}

We illustrate our findings in the following fictitious example of loan default prediction: given a bank loan, what is the probability that the customer will default, i.e.\ not repay the loan? A real-world example is presented in \S \ref{subsec:Simpson}.

Logistic regression is the main approach to default prediction \cite{calabrese2013modelling}; we will consider it in \S \ref{sec:LogReg}. Ridge regression is less common, but has the benefit of more interpretable feature importance; this is the setup for our main theoretical results in \S \ref{sec:mainRes}.

We take the simple setup of two binary features, gender and occupation group. Example counts of default outcomes by the four subpopulation groups are shown in the following contingency table:

\begin{figure}[H]
\begin{center}
\begin{tabular}{cc*{2}{S[table-format=2]}}
\toprule
& & \multicolumn{2}{c}{Default} \\
\cmidrule{3-4}
Gender & Occupation Group & {No} & {Yes}\\
\midrule
Female & A & 15 & 15\\
& B & 10 &  14\\ \addlinespace
Male & A & 16 &  5\\
& B & 27 &  8\\
\bottomrule
\end{tabular}
\end{center}
\end{figure}

Though an artificial example, we nevertheless explain its connection to actual default prediction practice. Regarding occupation groups, it is a common practice, especially in the face of smaller sample sizes, to reduce the cardinality of categorical values by binning them into a smaller number of statistically coherent groups. About gender, in the EU and United States, financial institutions are prohibited from explicitly using gender in making loan decisions \cite{gender-directive-eu}. Nevertheless, practitioners can and do use proxy variables to reconstruct gender even when absent from a dataset \cite{uk-report-gender-proxies}.

To obtain information about causes of default one might wish to perform a ridge regression on this dataset (see \S \ref{subsec:RidgeRegression} for a detailed description of the setup). Ridge regression depends on a choice of regularization parameter $c$. This choice is notoriously difficult, see e.g.\ \cite[\S 7]{hastie2009elements}. We show that a wrong choice can actually lead to dramatically wrong conclusions. Suppose one chooses $c=5$. Using the encoding described in \S \ref{subsec:Data} with $X_1=0$ denoting female, $X_2=1$ denoting the occupation category B etc., one obtains regression coefficients (see \eqref{equ:closedFormRegr} below)
\[
    \hat{\beta} \approx \left(0.023 \quad 0.331 \right)^T.
\]
For a classification output we compute a \emph{trend indicator} with respect to the first variable $X_1$, similar to the relative risk, as described in \S \ref{subsec:RelativeRisk}:
\[
    \T_{X_2 = j} = (0~j)\cdot \hat{\beta} - (1~j)\cdot \hat{\beta}\approx -2.31\%,
\]
The trend indicator being below zero suggests that in both occupation groups females are less likely to cause default events. However, this conclusion is at least questionable, if not wrong. For example, running ordinary linear regression with least squares minimization one obtains
\[
    \hat{\beta}_{MLS} = \left(-0.001 \quad 0.374 \right)^T,
\]
leading to a positive trend indicator $\T_{X_2 = j}\approx 0.1\%$, suggesting that females are more likely to cause default. This coincides with the trend in the dataset itself.\par
We have seen that a bad choice for the regularization parameter leads to a wrong conclusion in a binary classification. In fact, one would obtain the wrong conclusion for any $c > \frac{3}{13}\approx 0.23$. We call the interval $(\frac{3}{13},\infty)$ the pathological regularization regime $\A_2$ for this dataset. An informal definition follows, with a precise definition in \S \ref{subsec:PathRegReg}.

\begin{definition}[informal]
    A \emph{pathological regularization regime} for a given dataset and a choice of regularized regression model is a set of regularization parameters such that for any such parameter, the trend obtained from the regression is reversed compared to the ``true trend'' of the dataset.
\end{definition}

In our main result, Theorem \ref{thm:main}, we give an easily verifiable if-and-only-if condition for a dataset arising from a $2\times 2\times 2$ contingency table to have a non-empty pathological regularization regime for ridge regression and, if existent, precisely specify it. This result should be seen as a warning to data science practitioners: one should not choose a regularization parameter contained in the pathological regularization regime.\par

It should be noted that the case of $2\times 2\times 2$ contingency tables is the smallest case in which pathological regularization regimes occur. However, it is not restricted to this case, the phenomenon also appears for higher dimensional data. See \S \ref{subsec:largerTables} for a discussion on this.


\begin{figure}
    \centering
    \includegraphics[width=\textwidth]{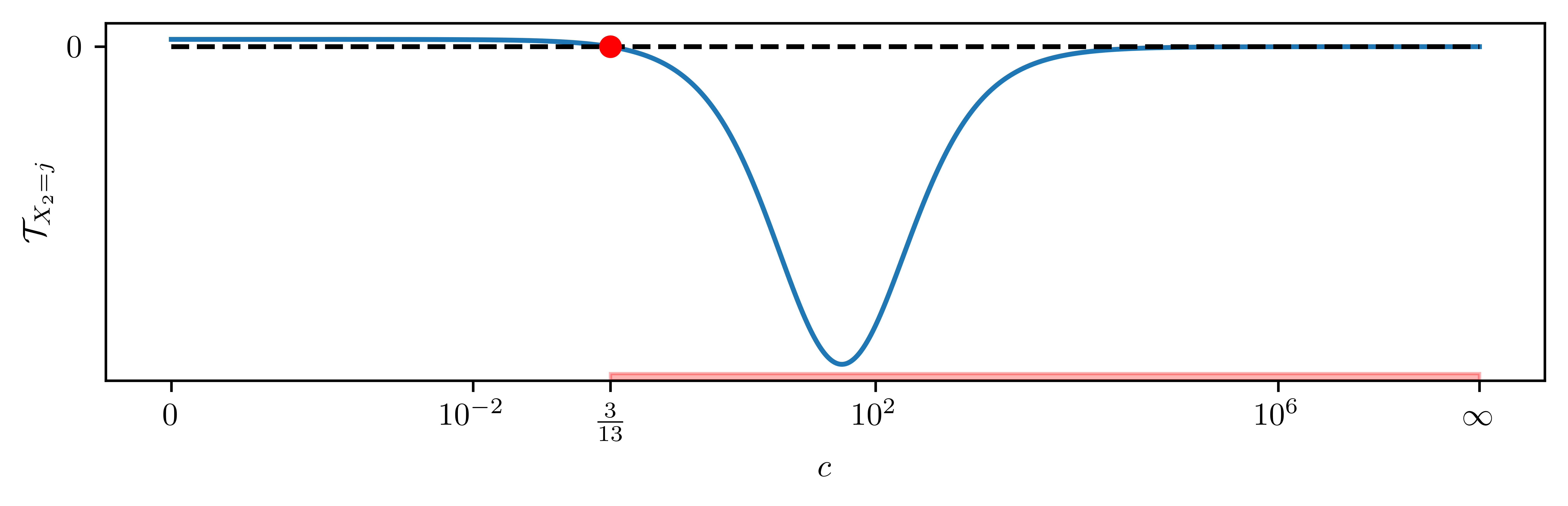}
    \vspace{-2em}
    \caption{The trend indicator from the credit score example with respect to the regularization parameter $c$. The interval marked in red is the pathological regularization regime.}
    \label{fig:exTrendInd}
\end{figure}

In \S \ref{subsec:Simpson} we study how likely it is to stumble upon such datasets with pathological regularization regimes in practice and draw connections to Simpson's paradox: Conjecture \ref{conj:conj} posits that every dataset exhibiting Simpson's paradox is pathological for logistic regression with intercept. \par

Our main contributions are summarized as follows:
\begin{itemize}
    \item definition of pathological regularization regimes (\S \ref{subsec:PathRegReg});
    \item proof of existence and characterization of pathological regularization regimes for ridge regression with and without intercept (\S \ref{subsec:ExistenceDescr} \& \S \ref{subsec:RemarkIntercept});
    \item study probability of occurrence of pathological regularization regimes and relations to Simpson's paradox (\S \ref{subsec:Simpson});
    \item conjecture that all Simpson datasets have pathological regularization regimes for logistic regression; this is supported by numerical evidence (\S \ref{sec:LogReg}).
\end{itemize}

\subsection{Prior Work}
Using regression for classification tasks is a well-studied subject, see for example \cite{huang2003linear,kocc2014application,naseem2010linear} for linear regression and \cite{dreiseitl2002logistic,lemon2003classification,ng2001discriminative} for logistic regression.
Some work can be found on how regularization influences the regression, e.g.\ \cite{salehi2019impact}. However, to the best of our knowledge, there does not exist any work on how regularization causes trend reversals in classification tasks similar to Simpson's paradox \cite{simpson1951interpretation}. In \cite{chen2009regression} the authors study a trend reversal between (unregularized) direct and reverse linear regression and relate it to Simpson's paradox. The blog post \cite{cprohm-causality} also demonstrates potential pathologies along regularization curves for ridge regression.\par
A key ingredient in our analysis is the study of non-monotonicity of regularization paths. While it is widely known that regularization paths need not be monotone, see e.g.\ \cite[\S 1.3]{van2015lecture}, we are not aware of any precise analysis when and where non-monotonicity occurs. \par
In the case of $\ell^1$-regularization, discontinuities of regularization paths or their derivatives have been studied, e.g.\ to find good regularization parameters \cite{schmidt2007learning} or to determine the complexity of learning algorithms \cite{bach2004computing}; note that in this paper we consider $\ell^2$-regularization and thus regularization paths will be smooth.

\section{Preliminaries}

We first explain how to encode a $2\times 2\times 2$ contingency table into a design matrix $X$ and a response variable $Y$ suitable for application of regression (\S \ref{subsec:Data}). Fundamentals of ridge regression are presented in \S \ref{subsec:RidgeRegression}, basics of logistic regression are recalled in \S \ref{subsec:LogRegPrelim}. Finally, in \S \ref{subsec:RelativeRisk} we introduce the trend indicator, converting the regression output into a score practical for classification.

\subsection{Data}
\label{subsec:Data}
In this work we are concerned with binary classification problems for two binary subpopulations. The input data is presented as a $2\times 2\times 2$ contingency table $D$. Let $N$ be the sample size, i.e.\ the sum of all entries in $D$. Then we can also represent the data $D$ as a tuple $(Y,X)$ where $Y$ is an $N$-dimensional vector, referred to as \emph{response variable}, and $X$ is an $N\times 2$ matrix, called \emph{design matrix}.\par
To pass from $D$ to $(Y,X)$ an \emph{encoding} is needed. In the forthcoming we will use the following encoding: let the entries of $D$ be labeled by $(i,j,k)\in\{0,1\}^3$. If the $(i,j,k)$th entry of $D$ is $d$, this will be encoded by $d$ entries of $i$ in $Y$ and $d$ rows of $(j,k)$ in $X$. An example for such an encoding would be as follows:
\vspace{1em}

\begin{minipage}{0.41\textwidth}
    \begin{flushleft}
        \centering
        \begin{tabular}{@{}cc|cc@{}}
        \toprule
        $X_1$ & $X_2$ & $Y=0$ & $Y=1$\\ \midrule
        0  & 0 & 1& 2\\
        0 & 1 & 0& 1\\
        1 & 0 & 3&  0\\
        1 & 1 & 1&  1\\\bottomrule
        \end{tabular}
    \end{flushleft}
\end{minipage}
\begin{minipage}{0.1\textwidth}
    \[
        \overset{\text{encoding}}{\leadsto}
    \]
\end{minipage}
\begin{minipage}{0.41\textwidth}
    \begin{align*}
        Y = & \begin{pmatrix}
            0 & 0 & 0 & 0 & 0 & 1 & 1 & 1 & 1
        \end{pmatrix}^T\\
        X= &\begin{pmatrix}
            0 & 1 & 1 & 1 & 1 & 0 & 0 & 0 & 1\\
            0 & 0 & 0 & 0 & 1 & 0 & 0 & 1 & 1
        \end{pmatrix}^T.
    \end{align*}
\end{minipage}
\vspace{1em}

Note, however, that all the results in this paper also hold true if we replace the $(0,1)$-encoding by an $(a,b)$-encoding for positive real numbers $a\neq b$.

\subsection{Ridge Regression}
\label{subsec:RidgeRegression}
Ridge regression is a version of linear regression introducing a penalization term for the parameters to prevent overfitting \cite[\S 3.4.1]{hastie2009elements}. Concretely, assume that the design matrix $X$ is of size $N\times p$ and $Y$ is an $N$-dimensional response variable; the \emph{regression estimator} is a $p$-dimensional vector $\hat{\beta}$ solving the optimization problem
\begin{equation}
    \label{equ:optimizationRR}
    \hat{\beta}(c) = \underset{\beta\in\R^p}{\argmin}~ ||Y - X\beta||^2_2 + c||\beta||^2_2.
\end{equation}
Here, $c\in\R_{>0}$ is the \emph{regularization parameter} controlling the strength of the regularization. In the limit case $c\rightarrow 0$, ridge regression becomes ordinary linear regression. There exists the following well-known closed form solution for $\hat{\beta}$, see e.g.\ \cite[(3.44)]{hastie2009elements}.
\begin{proposition}
    The ridge regression estimator is given by
    \begin{equation}
        \label{equ:closedFormRegr}
         \hat{\beta}(c) = \left(X^TX + c\id_p \right)^{-1}X^T Y,
    \end{equation}
    where $\id_p$ denotes the $p\times p$ identity matrix.
\end{proposition}
The following concept is of central importance to this paper.
\begin{definition}
    The \emph{regularization paths} for a ridge regression with design matrix $X$ and response variable $Y$ are the functions $\hat{\beta}_i\colon \R_{>0}\rightarrow\R,~c\mapsto \hat{\beta}_i(c)$ for $i=1,\dots,p$.
\end{definition}
The limit behavior of regularization paths is well-known, see e.g.\ \cite[\S 3.4.1]{hastie2009elements}.
\begin{lemma}
    The limits of the regularization paths are given by
    \label{lem:limitPaths}
    \[
        \lim_{c \downarrow 0} \hat{\beta}(c) = \hat{\beta}_{MLS} \quad \text{and} \quad \lim_{c\rightarrow\infty} \hat{\beta}(c) = 0,
    \]
    where $\hat{\beta}_{MLS}$ is the minimum least squares estimator, the sum-of-squares minimizer of minimal length, i.e.\
    \[\hat{\beta}_{MLS} = \argmin_{\beta\in\R^p} ||Y-X\beta||^2_2 \text{ and } ||\hat{\beta}_{MLS}||^2_2 < ||\beta||^2_2
    \]
    for all $\beta$ with $||Y-X\beta||^2_2 = ||Y-X\hat{\beta}_{MLS}||^2_2$.
\end{lemma}

\begin{figure}
    \centering
    \includegraphics[width=0.6\textwidth]{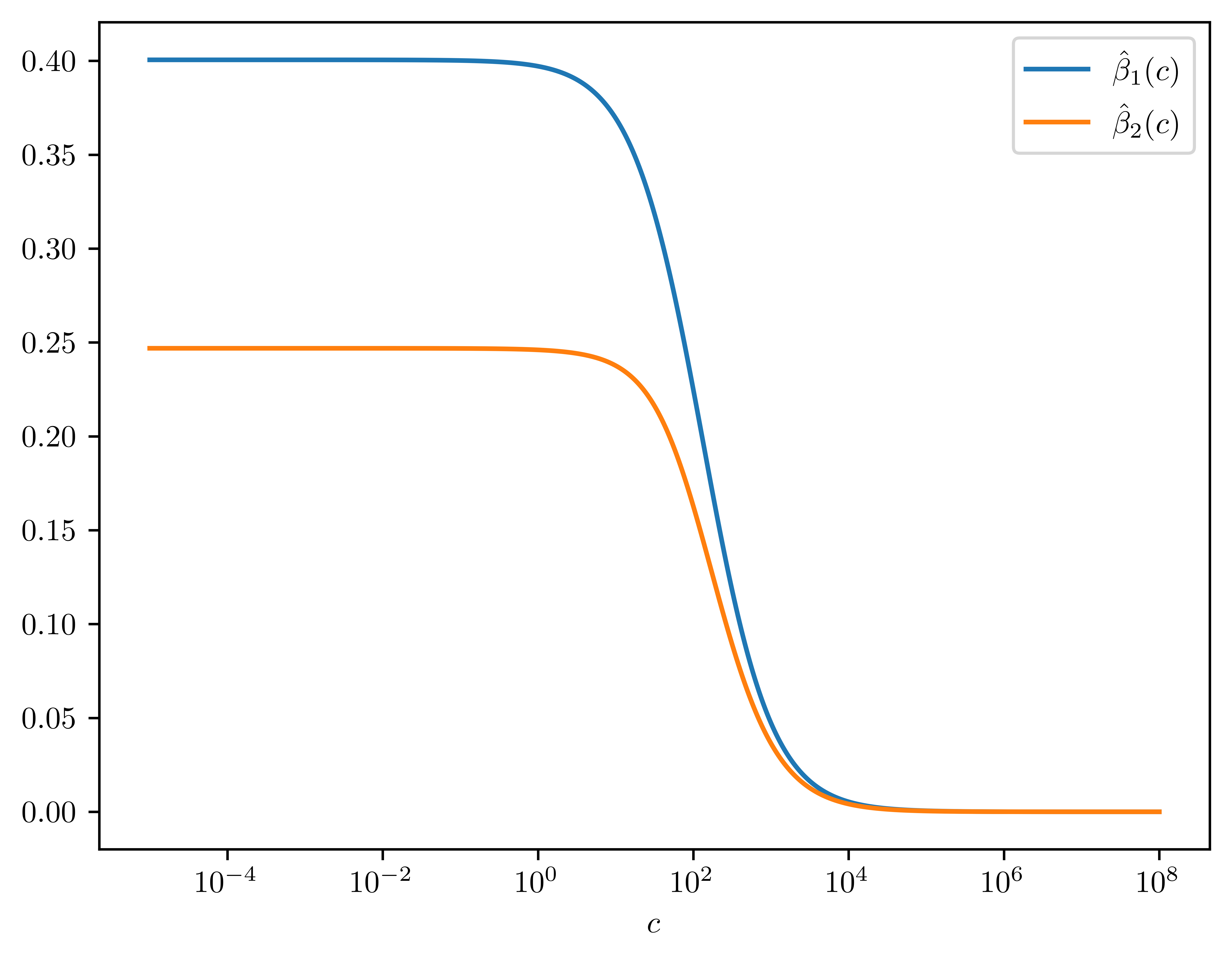}
    \caption{Two typical examples of regularization paths. In general, regularization paths need not be monotone; this leads to pathological regularization regimes.}
    \label{fig:ragPathsUsual}
\end{figure}

Oftentimes, regularization paths have the shape as the ones displayed in Figure \ref{fig:ragPathsUsual} (or mirrored along the $c$-axis). However, it is \emph{not} always the case that regularization paths are monotone. This non-monotonicity is the reason behind the existence of pathological regularization regimes (see \S \ref{subsec:PathRegReg}).

One might introduce an \emph{intercept term} into the regression, as is for example done in the standard implementation of ridge regression in the \texttt{scikit-learn} Python library \cite{sklearnRidge}. This amounts to centralizing the data \cite[\S 3.4.1]{hastie2009elements}. Our main results also extend to ridge regression with intercept; we make this precise in \S \ref{subsec:RemarkIntercept}. For simplicity, if not otherwise mentioned, we use ridge regression without intercept.

\subsection{Logistic Regression}
\label{subsec:LogRegPrelim}
In \S \ref{sec:LogReg} we will consider logistic regression; therefore, let us briefly recall the basics. See for instance \cite[\S 3]{friedman2010regularization} for more details.\par
The logit model is of the form
\[
    f(X^{\prime}) = \frac{1}{1 + \exp(-(\beta_0 + \beta\cdot X^{\prime}))},
\]
where in our case $X^{\prime}$ is a 2-dimensional vector, $\beta$ is a 2-dimensional parameter and $\beta_0\in\R$ is the intercept term. For an $N\times 2$ input matrix $X$ and an $N$-dimensional response variable $Y$ as obtained in \S \ref{subsec:Data}, the intercept and parameter estimates $\hat{\beta}_0$ and $\hat{\beta}$ are given by the maximum likelihood estimate
\begin{align*}
    (\hat{\beta}_0(c), \hat{\beta}(c)) & = \argmin_{\beta_0\in\R,~\beta\in\R^2} -\frac{1}{N}\left( \sum_{j=1}^N Y_j \log\left( \frac{1}{1+\exp(-(\beta_0 + \beta\cdot X_j^T))} \right)\right. \\
    &\left. + (1-Y_j) \log\left(1 - \frac{1}{1+\exp(-(\beta_0 + \beta\cdot X_j^T))} \right) \right) + c||\beta||^2_2,
\end{align*}
where $X_j$ denotes the $j^{\text{th}}$ row of $X$. Here, we are using again an $\ell^2$-regularization scaled by $c$. Note that, contrary to ridge regression, there does not exist a general closed from solution for the parameter estimate.

\subsection{Trend Indicator}
\label{subsec:RelativeRisk}

To use regression models for binary classification, we interpret its outputs as a score indicative of an input vector $X$ belonging to target output $Y$. The trend indicator is the difference between the scores obtained from distinct binary inputs. In an application, one can infer from the trend indicator statements of the form ``subgroup A is more likely to cause Y than subgroup B''.\par
More concretely, let $(Y,X)$ be a dataset obtained from a $2\times 2\times 2$ contingency table, representing binary random variables $X_1$, $X_2$ and $Y$, as in \S \ref{subsec:Data}. For a regression model $f_{\beta}$ with weights $\beta$, let $\hat{\beta}$ be the weight estimate. For binary input events $(X_1 = j, X_2 = k)$, we obtain a score by evaluating $f_{\hat{\beta}}$ at the vector $(j~k)^T$ and denote it by $f_{\hat{\beta}}(X_1 = j, X_2 = k)$.

\begin{definition}
    The \emph{trend indicator} for the input event $X_i = j$ (for $i\in\{1,2\}$) is
    \begin{equation}
        \label{equ:trendIndicator}
        \T_{X_i=j} \coloneqq f_{\hat{\beta}}(X_i = j, X_{\bar{i}} = 0) - f_{\hat{\beta}}(X_i = j, X_{\bar{i}} = 1).
    \end{equation}
    Here, $\bar{i}$ denotes the value in $\{1,2\}$ different from $i$.
\end{definition}

\begin{remark}
    If the parameter estimate $\hat{\beta}$ depends on a regularization parameter $c$ we emphasize this dependency in the trend indicator by writing
    \[
        \T_{X_i=j}(c) = f_{\hat{\beta}(c)}(X_i = j, X_{\bar{i}} = 0) - f_{\hat{\beta}(c)}(X_i = j, X_{\bar{i}} = 1).
    \]
\end{remark}

\begin{remark}
    \label{rem:independenceOfJ}
    In the case of ridge regression, the trend indicator simplifies as follows:
    \begin{align*}
        \T_{X_1=j}(c) & = (j~0)^T\cdot \hat{\beta}(c) - (j~1)^T\cdot \hat{\beta}(c) = -\hat{\beta}_2(c) \\
        \T_{X_2=j}(c) & = (0~j)^T\cdot \hat{\beta}(c) - (1~j)^T\cdot \hat{\beta}(c) = -\hat{\beta}_1(c).
    \end{align*}
    In particular, the trend indicator does not depend on the specific input $j$, and we might simply write $\T_{i}$ instead of $\T_{X_i=j}$.
\end{remark}

\subsection{Pathological Regularization Regimes}
\label{subsec:PathRegReg}

In this subsection we define the main object of this paper, the pathological regularization regime. The idea is that for a hyperparameter choice in this regime there will occur a trend reversal between the trained model and the actual dataset. \par
Let $\hat{\beta}(c)$ be a model estimate with regularization parameter $c$; for the purpose of this paper, this will either be ridge or logistic regression with $\ell^2$-regularization scaled by $c$. In the limit case, $\lim_{c \downarrow 0} \hat{\beta}(c)$, the estimate approaches the model estimate without regularization, i.e.\ linear regression or ordinary logistic regression, confer Lemma \ref{lem:limitPaths}. The trend in this limit case, as obtained from the trend indicator \eqref{equ:trendIndicator}, is what we refer to as the \emph{true trend}. Note that, by consistency of the estimates for linear and logistic regression (see \cite[\S 3.3]{mood1950introduction}), for large enough datasets the true trend should coincide with the trend directly obtained from the dataset.

\begin{definition}
    Given a dataset $(Y,X)$ obtained from a $2\times 2\times 2$ contingency table, the \emph{pathological regularization regime} $\A_{i,j}$ for the outcome $X_i=j$ (for $i,j\in\{1,2\}$) is the (possibly empty) subset $\A_{i,j}\subset \R_{>0}$ such that for any $c\in\A_{i,j}$, the sign of the trend indicator $\sgn(\T_{X_i=j}(c))$ is reversed compared to the true trend $\sgn(\lim_{c^{\prime}\downarrow 0} \T_{X_i=j}(c^{\prime}))$.
\end{definition}

We say that the dataset $(Y,X)$ has a pathological regularization regime if there exist $i,j \in \{1,2\}$ such that $\mathcal{P}_{i,j}$ is non-empty. For ridge regression, the trend indicators do not depend on $j$ (see Remark \ref{rem:independenceOfJ}) and we omit it from notation by simply writing $\A_i$.

\section{Main Results}
\label{sec:mainRes}

We first give an if-and-only-if condition on the existence of a pathological regularization regime for ridge regression and, if existent, describe it precisely in \S \ref{subsec:ExistenceDescr}. In \S \ref{subsec:Simpson} we explain how the probability to observe pathological regularization regimes and their range change with the sample size and draw a connection to Simpson's paradox.

\subsection{Existence and Description of Pathological Regularization Regimes}
\label{subsec:ExistenceDescr}

The following Lemma is the crucial ingredient in proving Theorem \ref{thm:main}.

\begin{lemma}
    \label{lem:zeroRegPath}
    Let $X$ be an $N\times 2$ design matrix and let $Y$ be an $N$-dimensional response variable. The regularization path $\hat{\beta}_i$ for a ridge regression on $(Y,X)$ has a positive zero if and only if the inequality
    \begin{equation}
        \label{equ:InequPosRoot}
        \left(\sum_{j=1}^N x_{j,1}x_{j,2}\right)\left(\sum_{j=1}^N x_{j,\bar{i}}y_{j}\right) > \left(\sum_{j=1}^N x_{j,\bar{i}}^2\right)\left(\sum_{j=1}^N x_{j,i}y_j\right)
    \end{equation}
    is satisfied. In this case, the zero is given by
    \begin{equation}
        \label{equ:zeroDescr}
        \left(\sum_{j=1}^N x_{j,i}y_j\right)^{-1} \left[ \left(\sum_{j=1}^N x_{j,1}x_{j,2}\right)\left(\sum_{j=1}^N x_{j,\bar{i}}y_{j}\right) - \left(\sum_{j=1}^N x_{j,\bar{i}}^2\right)\left(\sum_{j=1}^N x_{j,i}y_j\right)\right],
    \end{equation}
    and, moreover, $\hat{\beta}_i$ has a unique positive critical point. Here, $\bar{i}$ denotes the value in $\{1,2\}$ different from $i$.
\end{lemma}

\begin{proof}
    To simplify notation, any sum will range over $j=1,2,\dots,N$. Expanding the expression \eqref{equ:closedFormRegr} for the parameter estimate, we get
    \[
        \hat{\beta}(c) = \frac{1}{D(c)} \begin{pmatrix}
            c\left(\sum x_{j,1}y_j\right) + \left(\sum x_{j,2}^2\right)\left(\sum x_{j,1}y_j\right)
            - \left(\sum x_{j,1}x_{j,2}\right)\left(\sum x_{j,2}y_j\right) \\
            c\left(\sum x_{j,2}y_j\right) + \left(\sum x_{j,1}^2\right)\left(\sum x_{j,2}y_j\right)
            - \left(\sum x_{j,1}x_{j,2}\right)\left(\sum x_{j,1}y_j\right) \\
        \end{pmatrix},
    \]
    where $D(c) = c^2 + c\left(\sum x_{j,1}^2 + x_{j,2}^2\right) + \left(\sum x_{j,1}^2\right)\left(\sum x_{j,2}^2\right) - \left(\sum x_{j,1}x_{j,2}\right)^2$.
    As all entries of $X$ and $Y$ are non-negative,
    we see that $\hat{\beta}_i(c)$ has a unique positive root if and only if \eqref{equ:InequPosRoot} is satisfied. Clearly, this root is given by \eqref{equ:zeroDescr}.
    Moreover, we find that $\frac{\partial \hat{\beta}_i(c)}{\partial c} = 0$ if and only if
    \begin{align*}
        & c^2 \left(- \sum x_{j,i}y_j \right)
        +  2c\left[\left(\sum x_{j,1}x_{j,2}\right)\left(\sum x_{j,\bar{i}}y_j\right) - \left(\sum x_{j,\bar{i}}^2\right)\left(\sum x_{j,i}y_j\right)\right] \\
        + & \left(\sum x_{j,i}y_j\right)\left[\left(\sum x_{j,1}^2\right)\left(\sum x_{j,2}^2\right)- \left(\sum x_{j,1}x_{j,2}\right)^2\right] \\
        - & \left(\sum x_{j,1}^2+x_{j,2}^2\right)\left[\left(\sum x_{j,1}x_{j,2}\right)\left(-\sum x_{j,\bar{i}}y_j\right) + \left(\sum x_{j,\bar{i}}^2\right)\left(\sum x_{j,i}y_j\right)\right] = 0.
    \end{align*}
    For non-negative data, the leading coefficient is always negative. The linear coefficient is positive
    if and only if \eqref{equ:InequPosRoot} is satisfied. In this case, the constant term is also positive, by the Cauchy--Schwarz inequality applied to the first
    constant term. Hence, by Descartes' Rule of Signs \cite{descartes1637}, this quadratic equation has precisely one
    positive root if and only if \eqref{equ:InequPosRoot} is satisfied which concludes the claim.
\end{proof}

\begin{figure}
    \centering
    \includegraphics[width=0.6\textwidth]{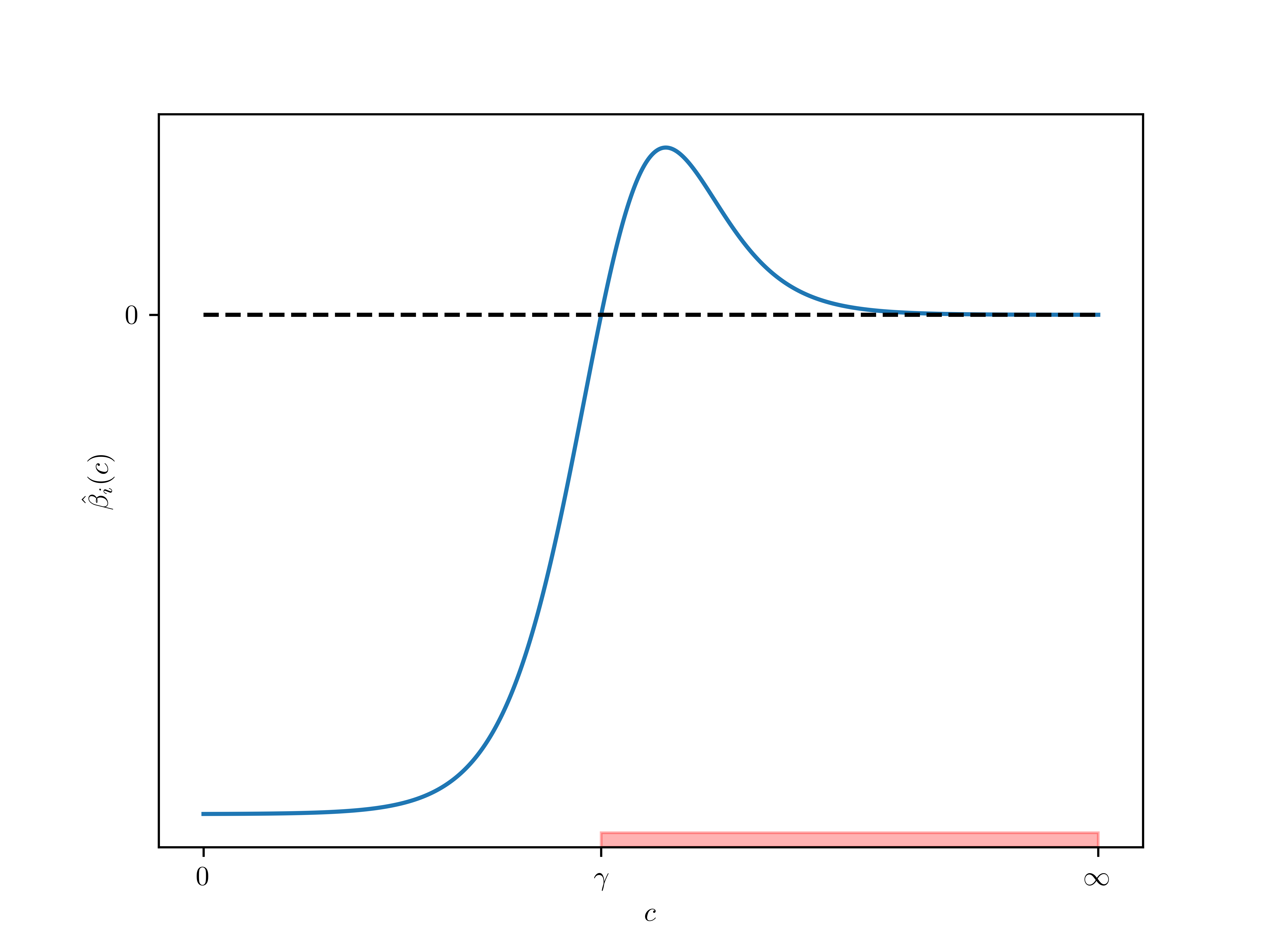}
    \caption{The shape of a regularization path satisfying inequality \eqref{equ:InequPosRoot} according to Lemmas \ref{lem:limitPaths} and \ref{lem:zeroRegPath}. The regime marked in red contributes to pathological behavior.}
    \label{fig:regPathWithZero}
\end{figure}

Figure \ref{fig:regPathWithZero} describes what a regularization path obtained from a dataset satisfying inequality \eqref{equ:InequPosRoot} heuristically looks like.

\begin{theorem}
    \label{thm:main}
    Binary classification ridge regression on data $(Y,X)$ has a pathological regularization regime if and only if there exists $i\in\{1,2\}$ such that
    \[
        \left(\sum_{j=1}^N x_{j,1}x_{j,2}\right)\left(\sum_{j=1}^N x_{j,\bar{i}}y_{j}\right) > \left(\sum_{j=1}^N x_{j,\bar{i}}^2\right)\left(\sum_{j=1}^N x_{j,i}y_j\right).
    \]
    In this case, the pathological regularization regime $\A_{\bar{i}}$ is given by $\A_{\bar{i}} = (\gamma,\infty)$ with
    \begin{gather*}
        \gamma = \left(\sum_{j=1}^N x_{j,i}y_j\right)^{-1} \left[ \left(\sum_{j=1}^N x_{j,1}x_{j,2}\right)\left(\sum_{j=1}^N x_{j,\bar{i}}y_{j}\right) - \left(\sum_{j=1}^N x_{j,\bar{i}}^2\right)\left(\sum_{j=1}^N x_{j,i}y_j\right)\right].
    \end{gather*}
\end{theorem}

\begin{proof}
    The trend indicator $\T_{\bar{i}}(c) = -\hat{\beta}_i(c)$ exhibits a trend reversal if and only if there exists a regularization parameter regime $\A_{\bar{i}}\subset \R_{>0}$ such that if $c\in \A_{\bar{i}}$ then
    \begin{equation}
        \label{equ:signChange}
        \sgn(\hat{\beta}_i(c)) = -\sgn\left(\lim_{c^{\prime}\downarrow 0} \hat{\beta}_i(c^{\prime})\right),
    \end{equation}
    where $\sgn$ is the sign function. This implies the existence of a positive root $\gamma$ of $\hat{\beta}_i(c)$. By Lemma \ref{lem:zeroRegPath}, such a zero implies the existence of a unique critical point. From the limit behavior in Lemma \ref{lem:limitPaths} we conclude that this critical point must be contained in $(\gamma, \infty)$. Hence, there must be a unique sign change of $\hat{\beta}_i$. Therefore, the sign change \eqref{equ:signChange} is equivalent to the existence of a positive root of $\hat{\beta}_i$. The rest of the statement is then a direct consequence of Lemma \ref{lem:zeroRegPath}.
\end{proof}

\subsection{Ridge Regression with Intercept}
\label{subsec:RemarkIntercept}
It is common practice to introduce an \emph{intercept} $\beta_0$ into the regression so that the model becomes $Y = X\beta + (\beta_0~\beta_0\cdots \beta_0)^T$. Typically, this intercept is not penalized in the ridge regression. The estimate can then be expressed as follows, see \cite[\S 3.4.1]{hastie2009elements}: the intercept estimate is $\hat{\beta}_0 = \bar{y} = \frac{1}{N}\sum_{j=1}^N y_j$. The remaining parameters get estimated via the closed form expression for ridge regression without intercept \eqref{equ:closedFormRegr}, replacing the entries of $X$ with centered inputs $x_{j,i} \mapsto \Tilde{x}_{j,i} \coloneqq x_{j,i} - \sum_{j^{\prime}=1}^N x_{j^{\prime},i}$. Note that no longer all $\Tilde{x}_{j,i}$ need to be positive. However, we can modify Theorem \ref{thm:main} as follows.

\begin{thmbis}{thm:main}
    Binary classification ridge regression \emph{with intercept} on data $(Y,X)$ has a pathological regularization regime if and only if there exists $i \in \{1,2\}$ such that
    \begin{gather*}
        \left(\sum_{j=1}^N \Tilde{x}_{j,i}y_j\right)^{-1}\left(\sum_{j=1}^N \Tilde{x}_{j,1}\Tilde{x}_{j,2}\right)\left(\sum_{j=1}^N \Tilde{x}_{j,\bar{i}}y_{j}\right) > \left(\sum_{j=1}^N \Tilde{x}_{j,i}y_j\right)^{-1}\left(\sum_{j=1}^N \Tilde{x}_{j,\bar{i}}^2\right)\left(\sum_{j=1}^N \Tilde{x}_{j,i}y_j\right).
    \end{gather*}
    In this case, the pathological regularization regime $\A_{\bar{i}}$ is given by $\A_{\bar{i}} = (\gamma,\infty)$ with
    \begin{gather*}
        \gamma = \left(\sum_{j=1}^N \Tilde{x}_{j,i}y_j\right)^{-1} \left[ \left(\sum_{j=1}^N \Tilde{x}_{j,1}\Tilde{x}_{j,2}\right)\left(\sum_{j=1}^N \Tilde{x}_{j,\bar{i}}y_{j}\right) - \left(\sum_{j=1}^N \Tilde{x}_{j,\bar{i}}^2\right)\left(\sum_{j=1}^N \Tilde{x}_{j,i}y_j\right)\right].
    \end{gather*}
    Here, $\Tilde{x}_{j,i}$ denotes the centered entry of the design matrix. (Note that the inequality above is equivalent to the inequality in Theorem \ref{thm:main} except for a possible sign change caused by the factor $\sum_{j=1}^N \Tilde{x}_{j,i}y_j$.)
\end{thmbis}

\begin{proof}
    Realizing that the trend indicator with intercept
    \begin{align*}
        \T_{X_1=j}(c) & = \left((j~0)^T\cdot \hat{\beta}(c) + \beta_0\right) - \left((j~1)^T\cdot \hat{\beta}(c) + \beta_0\right) = -\hat{\beta}_2(c) \\
        \T_{X_2=j}(c) & = \left( (0~j)^T\cdot \hat{\beta}(c) + \beta_0\right) - \left( (1~j)^T\cdot \hat{\beta}(c) + \beta_0\right) = -\hat{\beta}_1(c).
    \end{align*}
    is equal to the trend indicator without intercept, the proof is analogous to the proofs of Lemma \ref{lem:zeroRegPath} and Theorem \ref{thm:main}.
\end{proof}

\subsection{Sample Size and Simpson's Paradox}
\label{subsec:Simpson}

Two natural questions to consider are how likely it is to ``stumble'' over a dataset having a pathological regularization regime and whether this depends on the sample size $N$. We address these questions in this section. Moreover, we show how the pathological regularization regime changes with increasing sample size.\par

First note the following result.

\begin{proposition}
    \label{prop:arr-to-zero-bernoulli}
    If $X$ and $Y$ are matrices of size $N\times 2$ resp.\ $N\times 1$ with entries sampled from $\mathrm{Ber}(0.5)$ then the probability that the dataset $(Y,X)$ has a pathological regularization regime for ridge regression converges to zero as $N \rightarrow\infty$.
\end{proposition}

\begin{proof}
    Note that the expected value of $\sum_{j=1}^N x_{j,1}x_{j,2}$ is $N/4$ whereas the expected value of $\sum_{j=1}^N x_{j,\bar{i}}^2$ is $N/2$. Then the result follows from Theorem \ref{thm:main} and the law of large numbers.
\end{proof}

However, sampling the entries of $X$ and $Y$ randomly from $\{0,1\}$ is \emph{not} equivalent to sampling uniformly from $2\times 2\times 2$ contingency tables with a fixed sample size. We are now considering this setup.\par
Let
\[
    \mathcal{D}_N\coloneqq \{(d_{ijk})_{ijk} \in \Z_{\geq 0}^{2\times 2\times 2} \mid d_{+++} = N\}
\]
be the set of all $2\times 2\times 2$ contingency tables with sample size $N$; here, ``+'' denotes summation over all possible values for the respective index, e.g.\ $d_{0+1} = d_{001} + d_{011}$.

\begin{theorem}
    \label{thm:arr-to-constant-sample-contingencies}
    Let $D$ be sampled uniformly from $\mathcal{D}_N$ and let $(Y,X)$ be the response variable and design matrix obtained from $D$. Then the probability that $(Y,X)$ has a pathological regularization regime for ridge regression converges to a constant $\delta > 0$ as $N\rightarrow \infty$.
\end{theorem}

\begin{figure}
    \centering
    \includegraphics[width=0.6\textwidth]{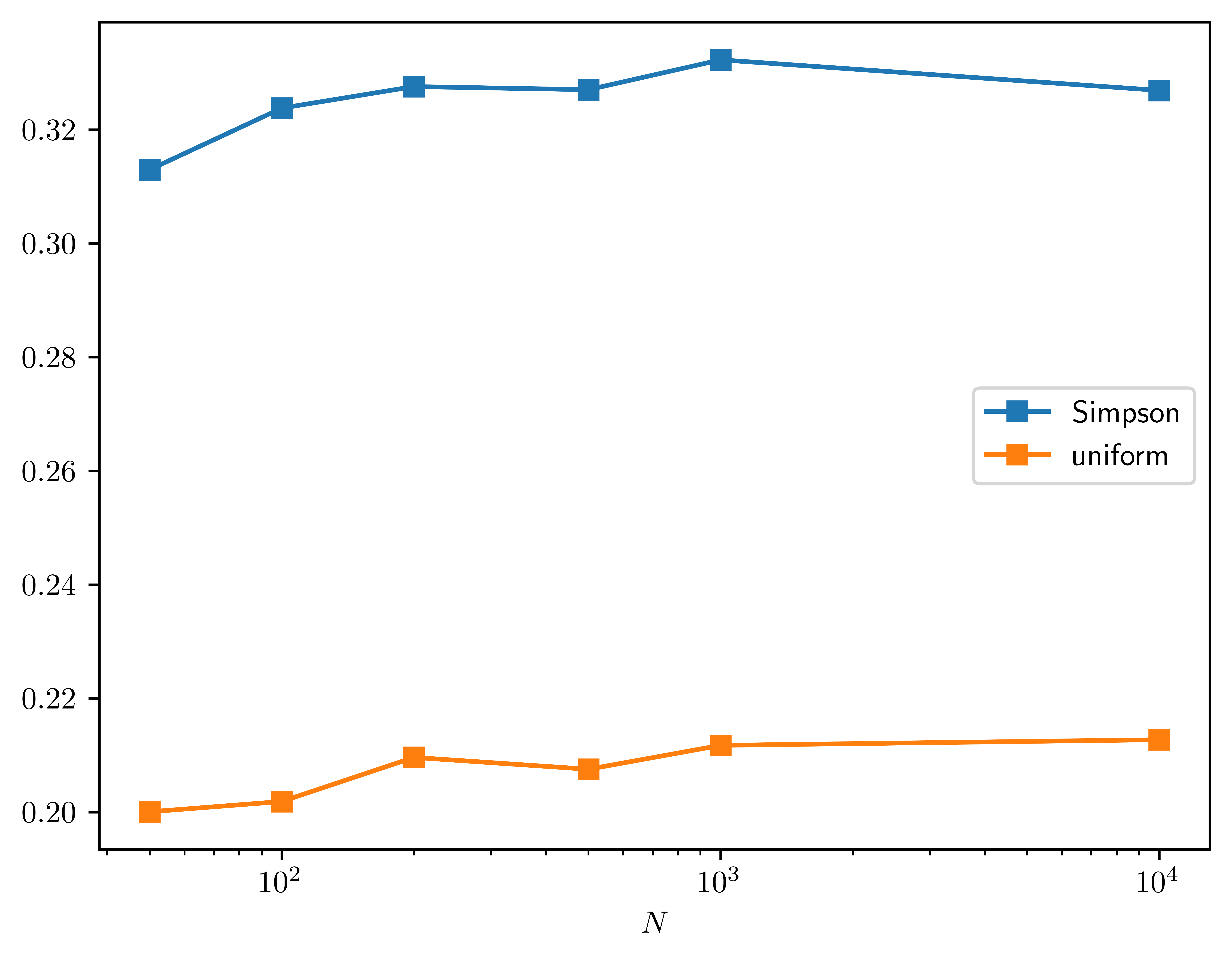}
    \caption{Proportion of datasets having pathological regularization regimes for different sample sizes if uniformly sampled from arbitrary contingency tables (orange) or from those exhibiting Simpson's paradox (blue).}
    \label{fig:ratioSampleSizeSimpson}
\end{figure}

\begin{remark}
    We numerically compute $\delta \approx 0.21$ via Monte Carlo simulation, see Figure \ref{fig:ratioSampleSizeSimpson}.
\end{remark}

\begin{proof}
    W.l.o.g.\ we show the statement for the pathological regularization regime $\A_2$. Rewriting inequality \eqref{equ:InequPosRoot} in terms of $D$ (for $i=1$) leads to
    \begin{equation*}
        d_{+11}d_{1+1} > d_{++1}d_{11+}.
    \end{equation*}
    Let $\bar{d}_{ijk} = \frac{d_{ijk}}{N}$ denote the normalized entries of $D$. For $N\rightarrow\infty$, the sampling process for $\bar{d}_{ijk}$ weakly converges to a Dirichlet $(1,1,\dots,1)$ distribution on the simplex (see \cite{during2017stylised} for how to make this precise) and hence amounts to uniformly sampling a probability distribution $(p_{ijk})_{ijk}$ from the probability simplex
    \[
        \Delta = \{(p_{ijk})_{ijk} \in \R_{\geq 0}^8 \mid p_{+++}=1\} \subset \R^8.
    \]
    The quadratic equation $p_{+11}p_{1+1} - p_{++1}p_{11+} = 0$ cuts out a hypersurface inside $\R^8$; on one side of this hypersurface there will be all points $(p_{ijk})_{ijk}$ satisfying $p_{+11}p_{1+1} > p_{++1}p_{11+}$. We need to show that the intersection of this region with the probability simplex $\Delta$ is full-dimensional, i.e.\ has positive Lebesgue measure. But this is clear as for example a small enough ball around the empirical probability distribution from the introductory example in \S \ref{subsec:Example} will lie in this intersection. This shows $\delta>0$.
\end{proof}

\begin{figure}
    \centering
    \includegraphics[width=0.6\textwidth]{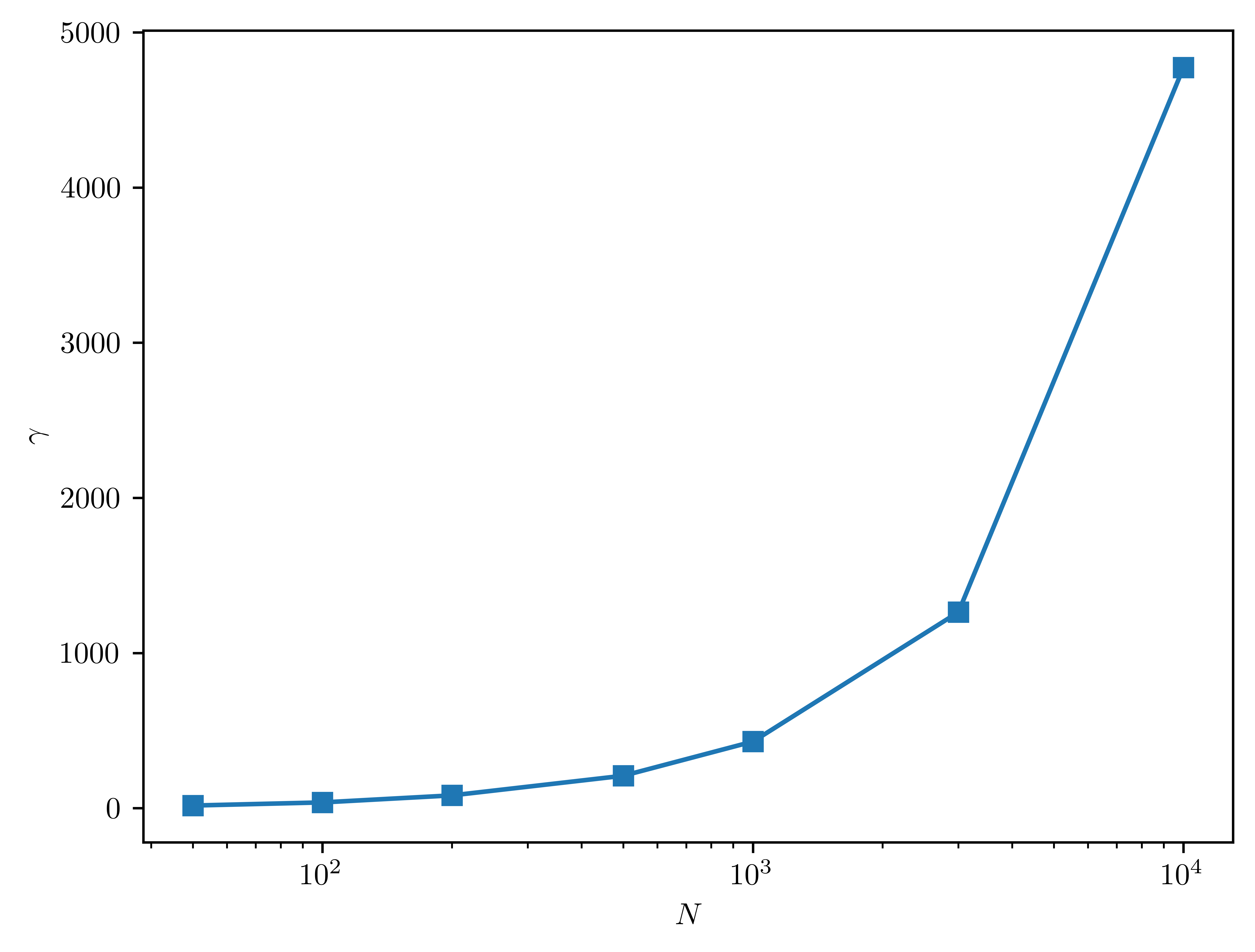}
    \caption{The average value of $\gamma$ among $10^4$ datasets uniformly drawn from $\mathcal{D}_N$ having a pathological regularization regime.}
    \label{fig:czero}
\end{figure}

While the probability to draw a dataset having a pathological regularization regime is constant with respect to the sample size, the regime typically shifts to the right, i.e. datasets with large sample sizes typically only show pathological regularization behavior for large regularization parameter, thus making it less likely one would accidentally choose such a bad parameter. In Figure \ref{fig:czero} below we show the average value of $\gamma$, the left bound of the pathological regularization regime $\A_i = (\gamma,\infty)$, for different sample sizes. \par

In practical applications, the general heuristic is to prefer stronger regularization for smaller datasets, as the risk of overfitting is higher. Combined with the above observations, the danger of choosing a pathological regularization parameter is especially high for smaller datasets.

Interestingly, if one only samples from datasets exhibiting \emph{Simpson's paradox} the probability of obtaining a dataset having a pathological regularization regime is increased to $\sim 32\%$. This might seem surprising as the probability of sampling a probability distribution exhibiting Simpson's paradox is less than $\frac{1}{12}$, see \cite{hadjicostas1998asymptotic}.\footnote{In fact, a Monte Carlo simulation shows that the probability is about $1.66\% $.} However, it is well-known that datasets exhibiting Simpson's paradox appear frequently in practice, e.g.\ \cite{ABRAMSON19921480,Bickel398,norton2015simpson,wagner1982simpson}, thereby providing a natural source for datasets with pathological regularization regimes. Let us recall some basics of Simpson's paradox. \par
Let $p_{ijk} = \mathbb{P}(Y=i, X_1=j, X_2=k)$ be the joint probability distribution on three binary random variables $Y,X_1$ and $X_2$. We say that $(p_{ijk})_{ijk}$ exhibits Simpson's paradox (or simply, is Simpson) if there is a trend reversal in the $Y$ variable between all subpopulations and the overall population. More formally, $(p_{ijk})_{ijk}$ is Simpson if one of the following two sets of inequalities is satisfied:
\newline
\begin{minipage}{0.49\textwidth}
    \vspace{1em}
    \begin{equation*}
        \left\{
        \begin{aligned}
            p_{101}p_{+00} - p_{100}p_{+01} &< 0\\
            p_{111}p_{+10} - p_{110}p_{+11} &< 0\\
            p_{1+1}p_{++0} - p_{1+0}p_{++1} &> 0\\
        \end{aligned}
        \right.
    \end{equation*}
    \vspace{1em}
\end{minipage}\hfill
\begin{minipage}{0.49\textwidth}
    \vspace{1em}
    \begin{equation*}
        \label{equ:Simpson}
        \left\{
        \begin{aligned}
            p_{101}p_{+00} - p_{100}p_{+01} &> 0\\
            p_{111}p_{+10} - p_{110}p_{+11} &> 0\\
            p_{1+1}p_{++0} - p_{1+0}p_{++1} &< 0\\
        \end{aligned}
        \right.
    \end{equation*}
    \vspace{1em}
\end{minipage}

We say that a dataset $(Y,X)$ exhibits Simpson's paradox if the corresponding empirical probabilities satisfy either of the sets of inequalities above.\par
By sampling datasets $(Y,X)$ with different sample-sizes $N$ with uniformly distributed entries from $\{0,1\}$ and checking the Simpson inequalities above, we sample $10^5$ datasets exhibiting Simpson's paradox for each sample size under consideration. For each of these datasets we apply Theorem \ref{thm:main} to check for the existence of a pathological regularization regime and observe the following (also see Figure \ref{fig:ratioSampleSizeSimpson}).

\begin{observation}
    Among datasets exhibiting Simpson's paradox, the proportion of those having a pathological regularization regime stays constant at $\sim 32\%$ regardless of the sample size $N$.
\end{observation}

Hence, datasets exhibiting Simpson's paradox provide a natural source for datasets with pathological regularization regimes. As it is commonly known that those datasets appear in practice, this should be seen as a warning that many real-world datasets might have pathological regularization regimes. We conclude this section with the following famous example taken from \cite{radelet1981racial}.\par
The data summarizes death sentences in Florida counties between 1976 and 1977 depending on the race of the defendant and the victim.

\begin{figure}[H]
\begin{center}
\begin{tabular}{cc*{1}{w{r}{0.8cm}}{S[table-format=2]}}
\toprule
& & \multicolumn{2}{c}{Death Penalty} \\
\cmidrule{3-4}
Victim & Defendant & {No} & {Yes}\\
\midrule
White & White & 132 & 19 \\
& Black & 52 & 11 \\ \addlinespace
Black & White & 9 &  0\\
& Black & 97 &  6\\
\bottomrule
\end{tabular}
\quad
\begin{tabular}{c*{1}{w{r}{0.8cm}}{S[table-format=2]}}
\toprule
& \multicolumn{2}{c}{Death Penalty} \\
\cmidrule{2-3}
Defendant & {No} & {Yes}\\
\midrule
White & 141 & 19 \\ \addlinespace
Black & 149 & 17 \\
\bottomrule
\end{tabular}
\end{center}
\end{figure}

For any race of the victim, white defendants are less likely to receive the death penalty (left table). However, when aggregated over the victims, white defendants are more likely to receive the death penalty. Hence, this is an instance of Simpson's paradox. Indeed, this dataset also has a pathological regularization regime: from Theorem \ref{thm:main} we obtain that $\A_2 = (125 \frac{5}{6}, \infty)$.

\subsection{Beyond $2\times 2\times 2$ Contingency Tables}
\label{subsec:largerTables}

We emphasize that the consideration of $2\times 2\times 2$ contingency tables is merely for illustrative purposes and that pathological regularization regimes also exist in higher dimensional settings. \par
More concretely, consider a $p\times 2\times 2$ contingency table $D$. This encodes the outcome of $p$ binary random variables $X_1,X_2,\dots,X_p$ and a binary random variable $Y$. From ridge regression, one now obtains $p$ regularization paths $\hat{\beta}_1,\hat{\beta}_2,\dots, \hat{\beta}_p$. Fix $i\in \{1,\dots,p\}$; it is now possible to define a trend indicator as follows: for $j\in J := \{1,\dots,p\}\setminus \left\{ i \right\}$, specify outcomes $X_j = x_j$, where $x_j\in\left\{ 0,1 \right\}$. Then, for a regression model $f_{\beta}$ with weights $\beta$ and weight estimate $\hat{\beta}$, we define
\[
    \T_{\left\{ X_j=x_j \right\}_{j\in J}}(c) := f_{\hat{\beta}(c)}\left( \left\{ X_j = x_j \right\}_{j\in J}, X_i = 0 \right) - f_{\hat{\beta}(c)}\left( \left\{ X_j = x_j \right\}_{j\in J}, X_i = 1 \right).
\]
Analogously to Remark \ref{rem:independenceOfJ}, we notice that in the case of ridge regression, the trend indicator simplifies to $\T_{\left\{ X_j=x_j \right\}_{j\in J}}(c) = -\hat{\beta}_i(c)$. It is now easy to see from the closed form formula of the ridge regression estimator \eqref{equ:closedFormRegr} that pathological regularization regimes also exist for $p\times 2\times 2$ tables: take any pair $(Y,X)$ arising from a $2\times 2\times 2$ dataset $D$ having a pathological regularization regime according to Theorem \ref{thm:main}. W.l.o.g.\ assume that the regularization path $\hat{\beta}_1$ has a zero. Now append $p-2$ zero columns to the design matrix $X$, turning it into an $N\times p$ matrix; this is a design matrix arising from a $p\times 2\times 2$ contingency table $D^{\prime}$. The new regularization paths $\hat{\beta}$ and $\hat{\beta}$ are unchanged, hence a trend reversal still occurs with respect to the first random variable $X_1$, so $D^{\prime}$ has a pathological regularization regime. \par
\begin{figure}
    \centering
    \includegraphics[width=0.6\textwidth]{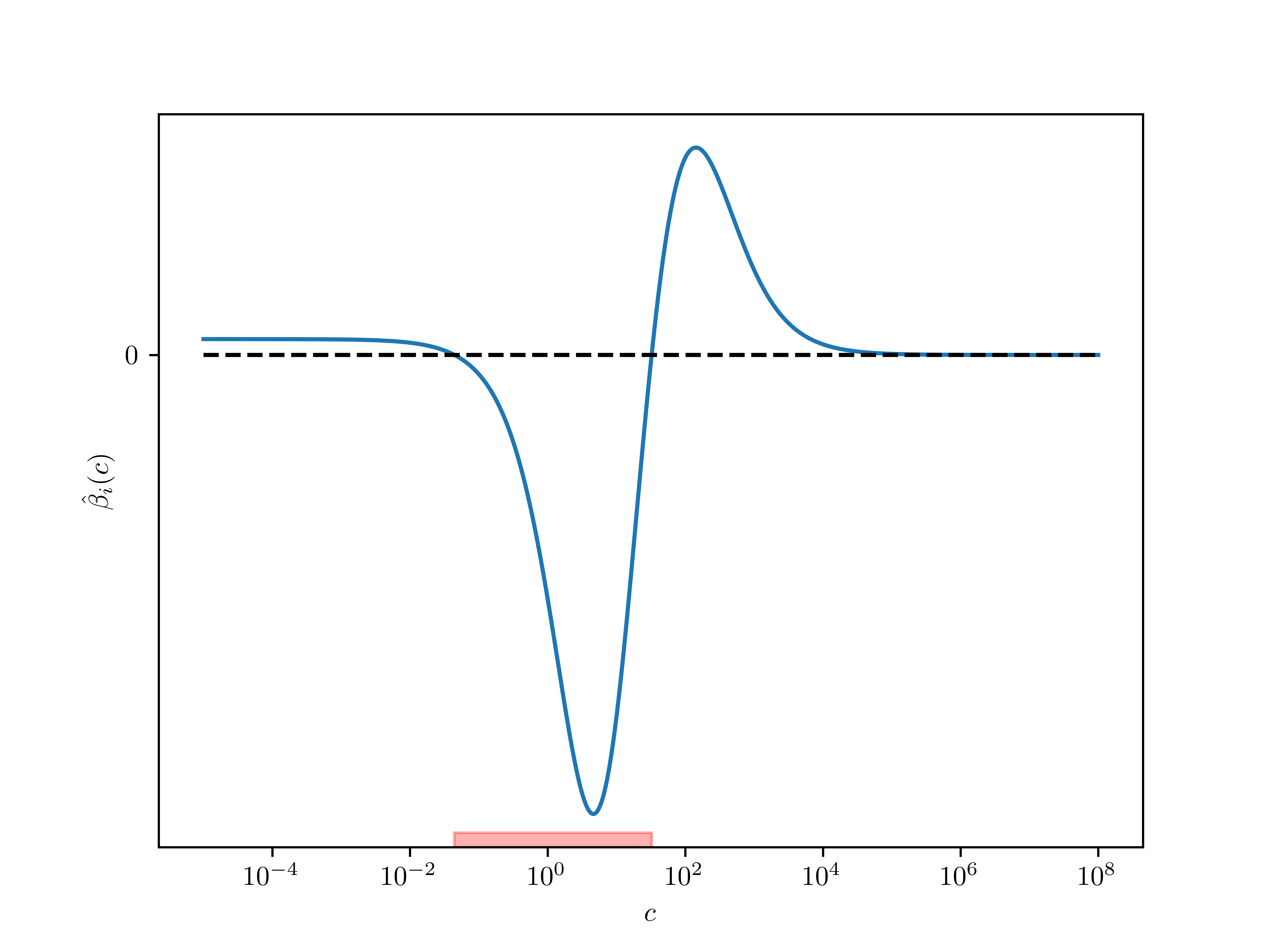}
    \caption{A regularization path arising from a $5\times 2\times 2$ contingency table; the pathological regularization regime (marked in red) is bounded.}
    \label{fig:beyond}
\end{figure}

 
In principle, it is still possible to derive inequalities in the entries of $X$ and $Y$ that provide necessary and sufficient conditions for a $p\times 2\times 2$ dataset having a pathological regularization regime, using Descartes' Rule of Signs \cite{descartes1637}. However, this would require a large case distinction and we do not find it instructive to list these inequalities here for $p>2$. It should be noted that the shape of pathological regularization regimes may change. In particular, they can be bounded or even disconnected. An example of the former in the case $p=5$ is presented in Figure \ref{fig:beyond}.

\section{Logistic Regression}
\label{sec:LogReg}

In this section we study pathological regularization regimes for logistic regression. Unlike ridge regression,
the logistic regression estimate does not admit closed form maximum likelihood solutions, so we mainly focus on numerical experiments.\par

\begin{figure}
    \centering
    \includegraphics[width=0.6\textwidth]{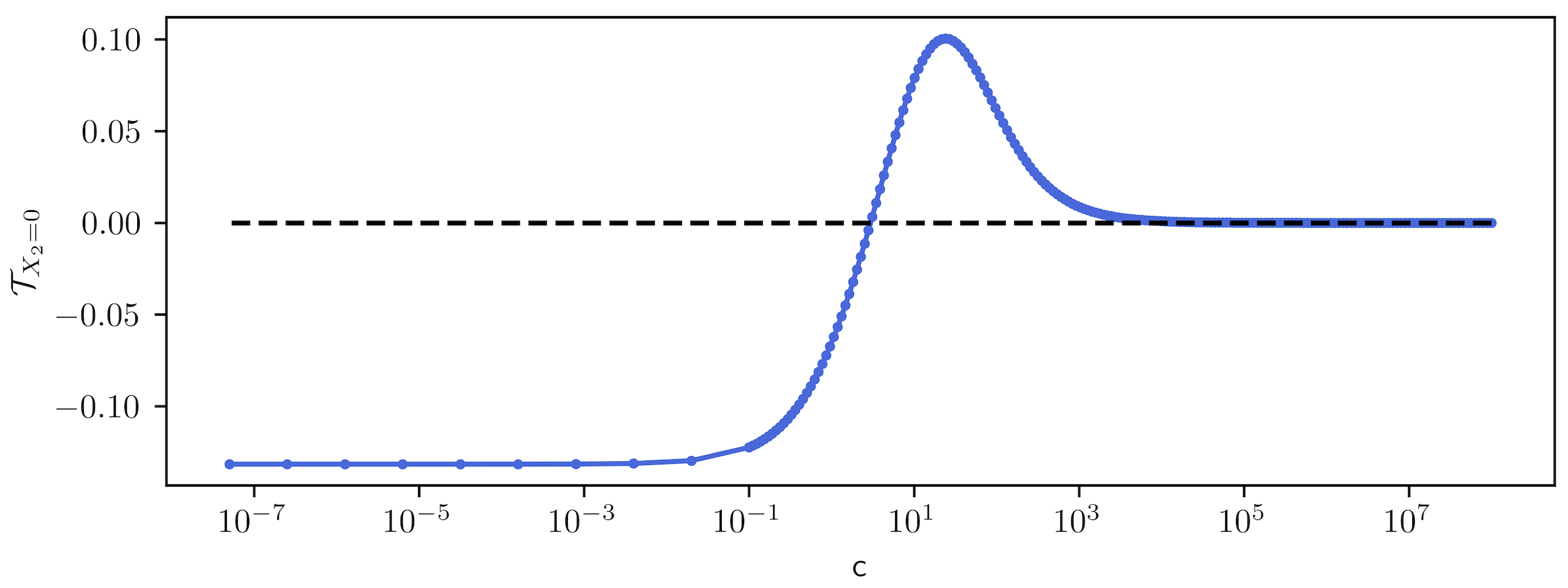}
    \caption{Trend curve for $X_2=0$ from a Simpson dataset with $N=400$ with intercept fitting.}
    \label{fig:logregpathologicalintercept}
\end{figure}

Indeed, pathological regularization regimes also exist for logistic regression. A typical example for such a trend reversal can be found in Figure \ref{fig:logregpathologicalintercept}.\par

It turns out that for logistic regression, datasets exhibiting Simpson's paradox play an even more peculiar role than for ridge regression.

\subsection{Pathological Ratios}

The goal of this section is to define and outline the estimation of \emph{pathological ratios} for logistic regression. The purpose of a pathological ratio is to quantify the percentage of a collection of datasets (e.g.\ $2\times 2\times 2$ Simpson datasets of sample-size 200) that have a non-trivial pathological regularization regime for some classification algorithm, e.g.\ logistic regression.

\begin{definition}
Given a collection of $2\times 2\times 2$ binary datasets $U$, and a classification model with parametrized regularization, the \emph{pathological ratio} of $U$ and the classification model $f$ is
the probability that a generic $(Y, X) \in U$ has a non-trivial pathological regularization regime for the fitted model $f_{\hat{\beta}}$.
\end{definition}

In the case of finite collections $U$, the pathological ratio is just the count of $(Y, X)$ that have at least one pathological regularization parameter divided by the cardinality of $U$.

We have already seen pathological ratios in Proposition \ref{prop:arr-to-zero-bernoulli}, where the statement can be reformulated to say that the pathological ratio of datasets $(Y, X)$ of size $N$ sampled uniformly from $\mathrm{Ber}(0.5)$ goes to 0 as $N \rightarrow \infty$ for ridge regression with intercept. Theorem \ref{thm:arr-to-constant-sample-contingencies} says that the pathological ratio stays constant over sample sizes $N$ for ridge regression without intercept for datasets derived from uniformly sampled contingency tables.

\subsection{Generating Simpson and non-Simpson Datasets}

To make conclusions about pathological regularization regimes for logistic regression, we require a way to produce statistically meaningful Simpson and non-Simpson datasets in bulk. Our basic approach is to use rejection sampling from the Dirichlet distribution, as introduced in \S \ref{subsec:Simpson}.

For a (dataset) sample size $N$, we want to uniformly generate $M$ Simpson and $M$ non-Simpson datasets, which we will then use to fit logistic regression for a range of regularization parameters to check if any of them is pathological. For such a regularization parameter $c$ exhibiting trend reversal we therefore have a proof that the given dataset has a non-trivial pathological regularization regime. See Algorithm \ref{alg:sample-data} for pseudocode.

\begin{algorithm}
    \caption{Sampling of $M$ Simpson datasets of size $N$}
    \label{alg:sample-data}
    \begin{algorithmic}[1]
        \STATE Initialize container \texttt{SIMPSON\_DATASETS} = $\{\}$
        \WHILE{$|\texttt{SIMPSON\_DATASETS}| < M$}
            \STATE Sample $p \in \Delta_7 \subset \mathbb{R}^8$ from the Dirichlet distribution
            \STATE Scale up $p$ to a contingency table $D$  by $D = \mathrm{round}(p \cdot N)$
            \IF{$\Sigma D \neq N$, i.e.\ rounding doesn't yield a dataset with $N$ records}
                \STATE Return to line 3
            \ENDIF
            \STATE Convert contingency table $D$ into a dataset $(Y, X)$ as per \S \ref{subsec:Data}
            \IF{$(Y, X)$ is Simpson}
                \STATE Append $(Y, X)$ to \texttt{SIMPSON\_DATASETS}
            \ENDIF
        \ENDWHILE
    \end{algorithmic}
\end{algorithm}

Our implementation of Algorithm \ref{alg:sample-data} can be found in the script \texttt{generate-data.py} in the repository \cite{prr-repo}.

\subsection{Numerical Computation of Pathological Ratios for Logistic Regression}

Once we have a collection of uniformly sampled Simpson and non-Simpson datasets across a range of sample-sizes $N$ per dataset, we use Monte Carlo sampling to estimate pathological ratios by fitting logistic regression to these datasets. To test for the existence of non-empty pathological regularization regimes, we run this logistic regression across a range of regularization parameters $c$. If one such parameter for a fixed subpopulation of the dataset $(Y, X)$ leads to trend reversal, then the dataset is considered pathological for logistic regression. In the current work, we always include the intercept term when fitting. Algorithm \ref{alg:fit-data} provides pseudocode for this approach.

For the sake of concreteness, we refer to subpopulations of our $2\times 2$ feature space as in the running example from \S \ref{subsec:Example}, namely the first variable $X_1$ gives the gender of a bank loan holder, while the second variable $X_2$ gives the occupation group, $A$ ($X_2 = 0$) or $B$ ($X_2 = 1$).

We estimate the pathological ratios by Monte Carlo simulation, as described in Algorithm \ref{alg:fit-data}.

\begin{algorithm}
    \caption{Logistic regression model fitting for pathological ratios}
    \label{alg:fit-data}
    \begin{algorithmic}[1]
        \STATE Given $M'$ datasets $U$, classification algorithm $f$, and collection of regularization parameters $c \in C$, where $0 \in C$ for no-regularization
        \STATE Initialize container \texttt{PATHOLOGICALS} = $\{\}$
        \FOR{each $(Y, X) \in U$}
            \FOR{each $c \in C$}
                \STATE Fit $f$ to $(Y, X)$ with regularization $c$, obtain $f_{\hat{\beta}}$
                \STATE Calculate trend indicator $\T_{X_2=j}(c)$ for each occupation value $j\in \{0, 1\}$
            \ENDFOR
            \IF{there exists $\sgn(\T_{X_2=j}(c)) \neq \sgn(\T_{X_2=j}(c))$ for some $c > 0$ (trend-reversal)}
                \STATE Append $(Y, X)$ to \texttt{PATHOLOGICALS}
            \ENDIF
        \ENDFOR
    \end{algorithmic}
\end{algorithm}

In our experiments, we chose based on theoretical and empirical grounds:

\begin{itemize}
    \item $\ell^2$ regularization to match the ridge regression sections above;
    \item sample sizes of $N=200$, 600 and 2400;
    \item the optimizer \texttt{newton-cholesky} of \texttt{scikit-learn} \cite{scikit-learn}; and
    \item regularization parameter grid with values $c$ chosen as
    \begin{itemize}
        \item 10 evenly log-spaced values between $10^{-8}$ and $10^{-1}$,
        \item 150 evenly log-spaced values between $10^{-1}$ and $10^{6}$,
        \item 40 evenly log-spaced values between $10^{6}$ and $10^{8}$.
    \end{itemize}
\end{itemize}

We will give more details and analyses for these choices in a follow-up paper, but remark briefly on the choice of numerical solver and regularization parameter grid. The optimizer \texttt{newton-cholesky} is suited to $\ell^2$ regularization with small- to medium-sized datasets. Moreover, it gave smoother trend-vs-regularization parameter curves than both the \texttt{scikit-learn} default \texttt{lbfgs} solver and the \texttt{sag} solver.

Regarding the choice of regularization parameter grid, we chose the stated grid by first considering an even finer log-spaced grid between $10^{-8}$ and $10^{8}$ for a collection of datasets, and then plotted the histogram distribution of the most pathological $c$ values (i.e.\ corresponding to the highest trend-reversal) per dataset. In our observations, most pathological $c$ values occurred between 1 and $10^4$. We put the largest (log-)concentration of grid points between $10^{-1}$ and $10^{6}$ to ensure coverage of the observed most pathological $c$'s, with a smaller number of grid points for very weak regularization and very strong regularization. Note that---given fixed hardware and parallelization---the run-time scales linearly with the number of regularization grid points. On the other hand, insufficient sampling of regularization values runs the risk of misclassifying pathological datasets / model combinations as non-pathological.

\subsection{Numerical Results and Conjecture}

Our main result about logistic regression is shown in Table \ref{table:main-num-result}, where we have taken $M=1250$ uniformly sampled Simpson and non-Simpson datasets for sample sizes $N=200$, 600 and 2400. For logistic regression with intercept, every Simpson dataset is pathological, and approximately 6.5\% of non-Simpson datasets are.

\begin{table}[ht]
    \centering
    \caption{Mean pathological ratios for Simpson and non-Simpson datasets for different sample sizes}
    \label{table:main-num-result}
    \scalebox{0.9}{\begin{tabular}{ccc}
    \toprule
    Sample size & Simpson mean pathological ratio & Non-Simpson mean pathological ratio \\
    \midrule
    200 & 1.0000 & 0.0632 \\
    600 & 1.0000 & 0.0680 \\
    2400 & 1.0000 & 0.0648 \\
    \bottomrule
    \end{tabular}}
\end{table}

Based on these numerical results, we conjecture that all Simpson datasets have a non-trivial pathological regularization regime.

\begin{conjecture}
    \label{conj:conj}
    All $2\times 2\times 2$ datasets strictly exhibiting Simpson's paradox are pathological for logistic regression with an intercept term.
\end{conjecture}

Additional evidence using larger Monte Carlo samples, as well as different numerical solvers will appear in the aforementioned follow-up paper. In particular, the confidence intervals for the non-Simpson pathological ratios in Table \ref{table:main-num-result} were not sufficient with $M=1250$ datasets to conjecture a constant pathological ratio, though the above plus not-yet-reported experiments confirm a ratio of approximately $6.5\%$.

\subsection{Pathological Datasets for Logistic Regression in Practice}

We have shown that non-trivial pathological regularization regimes can occur ``in-the-wild'' for both ridge and logistic regression by demonstrating families of datasets for which at least one regularization parameter $c$ exists resulting in trend reversal.

In practice, there are heuristics for finding good regularization values given a dataset, such as the one previously mentioned of regularizing more for small datasets and less for larger datasets. A more quantitative approach is to pick the regularization value $c$ that optimizes your classification problem's chosen success metric, e.g.\ accuracy.

The process of selecting an optimal $c$ given a dataset should not be carried out using the entire dataset, however, as this practice tends to cause overfitting, hence worsening generalization error; see e.g.\ \cite[\S 7]{hastie2009elements}.

In the case of smaller sample-sizes, the classic "train-validate-test" splitting of available data is often abandoned in favor of \emph{k-fold cross validation}, in which the available data is more or less chunked into $k$-subsets. The classification model is then trained on a choice of $k-1$ of these subsets for different $c$ values, with evaluation taking place using the hold-out $k$th subset. These different evaluations results are then combined to find an optimal $c$ for the k-fold split; see \cite[\S 7.10]{hastie2009elements}.

Perhaps the most worrisome pathological datasets for logistic regression would be if following the above ``best-practice'' procedure yields a regularization parameter $c$ in the pathological regime. In this section, we give an example of such a dataset, but save a more systematic study of this phenomenon for later work.

Consider the default dataset with 600 samples used in the trend-vs-regularization figure above, Figure \ref{fig:logregpathologicalintercept}. This data can be viewed and downloaded from \href{https://github.com/munichpavel/pathological-regularization-regimes/blob/main/data/pathological-default-for-x-validation.csv}{data/pathological-default-for-x-validation.csv} of the repository \cite{prr-repo}.

The dataset satisfies Simpson's paradox, with total population gender default trend of 0.45 (to two decimal places throughout), with subpopulation trends of $-0.23$ for both occupation groups.

We use scikit learn's \cite{scikit-learn} implementation of \href{https://scikit-learn.org/stable/modules/generated/sklearn.linear_model.LogisticRegressionCV.html}{logistic regression with cross validation}, changing the numerical solver to \texttt{newton-cholesky} as discussed above; see the code notebook \href{https://github.com/munichpavel/pathological-regularization-regimes/blob/main/notebooks/cross-validation.ipynb}{notebooks/cross-validation.ipynb} of \cite{prr-repo} for the precise code.

Recalling that the value of $X_2$ corresponds to our two occupation groups, the ``true trends'' obtained by no regularization are (rounding to three decimal places)
\[
    \T_{X_2=0}(c=0) = -0.196 \quad \text{and} \quad \T_{X_2=1}(c=0) = -0.234,
\]

whereas after fitting with cross validation as described above, the trends are
\[
    \T_{X_2=0}(c=\hat{c}) = 0.003 \quad \text{and} \quad \T_{X_2=1}(c=\hat{c}) = 0.003,
\]

resulting in a slight pathological trend reversal by following the best-practice of using $k$-fold cross validation.

It is straightforward to show that Simpson's paradox can only arise when subpopulation cardinalities differ, meaning that Simpson datasets are necessarily unbalanced. In such cases, one common approach is to weight the data samples, giving more weight to rare samples, and less to the frequent ones; see e.g.\ \cite{king2001logistic}. The effect of different mitigation approaches deserves more attention than we will give here, but we again use the default approach of scikit learn, namely specifying \texttt{class\_weight}='\texttt{balanced}' with \href{https://scikit-learn.org/stable/modules/generated/sklearn.linear_model.LogisticRegressionCV.html}{logistic regression with cross validation} \cite{scikit-learn}.

For this dataset, addressing the class imbalances actually worsens the trend reversal:
\[
    \T_{X_2=0}(c=\hat{c}) = 0.132 \quad \text{and} \quad \T_{X_2=1}(c=\hat{c}) = 0.117.
\]

This underscores that common best-practice procedures do not necessarily prevent pathological choices of regularization parameters.

\section{Conclusion and Outlook}

In this paper, we introduced the notion of a pathological regularization regime. For a regularization hyperparameter choice in this regime one observes a trend reversal in binary classification tasks between the dataset and a regression model. In the case of ridge regression we give a necessary and sufficient condition for the existence of such a regime and describe it explicitly. This can be used as a hands-on tool for a practitioner to avoid these bad regularization choices. We draw connections to Simpson's paradox by noticing that the probability of observing datasets having pathological regularization regimes is particularly high among datasets exhibiting Simpson's paradox. This also suggests that datasets having pathological regularization regimes appear frequently in real-world contexts. We show that pathological regularization regimes also exist for logistic regression and conjecture, based on numerical evidence, that all Simpson datasets are pathological for logistic regression with intercept term. Finally, we demonstrate that common best-practice procedures might not prevent pathological regularization parameter choices.\par
For future work it would be interesting to quantify pathological regularization regimes more precisely for logistic regression. This could lead to a deeper understanding of Conjecture \ref{conj:conj}. Moreover, the phenomenon should be investigated for different types of regularization, e.g.\ lasso techniques, and potentially be extended to generalized linear models.\par\bigskip

\textbf{Acknowledgments.} We are thankful to Alexander Kreiß, Guido Mont\'ufar and Martin Wahl for providing comments on an earlier draft of this article, and to Christopher Prohm for many valuable discussions.

\bibliographystyle{alpha}
\bibliography{bibliography}

\end{document}